%% file: acl_latex.tex
\pdfoutput=1

\documentclass[11pt]{article}
\usepackage{xspace}

\usepackage[final]{acl}

\usepackage{times}
\usepackage{latexsym}
\usepackage{amssymb}

\usepackage[T1]{fontenc}

\usepackage[utf8]{inputenc}

\usepackage{microtype}

\usepackage{inconsolata}

\usepackage{graphicx}

\usepackage{amsmath}
\usepackage{algorithm}
\usepackage{algorithmic}
\usepackage{newfloat}
\usepackage{listings}
\usepackage{multirow}
\usepackage{booktabs}
\usepackage{tcolorbox}
\input{definitions}
\newcommand{\our}{{QUITO-X}\xspace}
\title{QUITO-X: A New Perspective on Context Compression from the Information Bottleneck Theory}

\author{
 \textbf{Yihang Wang\textsuperscript{1,2}\thanks{Equal Contribution.}},
 \textbf{Xu Huang\textsuperscript{3}\footnotemark[1]},
 \textbf{Bowen Tian\textsuperscript{4}},
 \textbf{Yueyang Su\textsuperscript{1,2}\thanks{Corresponding author.}},
 \textbf{Lei Yu\textsuperscript{1,2}},
\\
 \textbf{Huaming Liao\textsuperscript{1,2}},
 \textbf{Yixing Fan\textsuperscript{1,2}},
 \textbf{Jiafeng Guo\textsuperscript{1,2}},
 \textbf{Xueqi Cheng\textsuperscript{1,2}}
\\
\\
 \textsuperscript{\rm 1}State Key Laboratory of AI Safety, ICT, CAS,\\
 \textsuperscript{\rm 2}University of Chinese Academy of Sciences,\\
 \textsuperscript{\rm 3}Peking University, \\
 \textsuperscript{\rm 4}Hong Kong University of Science and Technology (Guangzhou)
\\
 \small{
   \{\href{mailto:yihangwang1020@gmail.com}{yihangwang1020}, \href{mailto:ydove1031@gmail.com}{ydove1031}\}@gmail.com; suyueyang@ict.ac.cn
 }
}

\begin{document}
\maketitle
\begin{abstract}
Generative \acp{LLM} have achieved remarkable success in various industrial applications, owing to their promising In-Context Learning capabilities. However, the issue of long context in complex tasks poses a significant barrier to their wider adoption, manifested in two main aspects: (i) The excessively long context leads to high costs and inference delays. (ii) A substantial amount of task-irrelevant information introduced by long contexts exacerbates the "lost in the middle" problem.
Existing methods compress context by removing redundant tokens using metrics such as self-information or \ac{PPL}, which is inconsistent with the objective of retaining the most important tokens when conditioning on a given query. In this study, we introduce information bottleneck theory (IB) to model the problem, offering a novel perspective that thoroughly addresses the essential properties required for context compression. Additionally, we propose a cross-attention-based approach to approximate mutual information in IB, which can be flexibly replaced with suitable alternatives in different scenarios. Extensive experiments on four datasets demonstrate that our method achieves a 25\% increase in compression rate compared to the state-of-the-art, while maintaining question answering performance. In particular, the context compressed by our method even outperform the full context in some cases.
\end{abstract}

\section{Introduction}

\begin{figure}[t!]
\includegraphics[width=0.9\columnwidth]{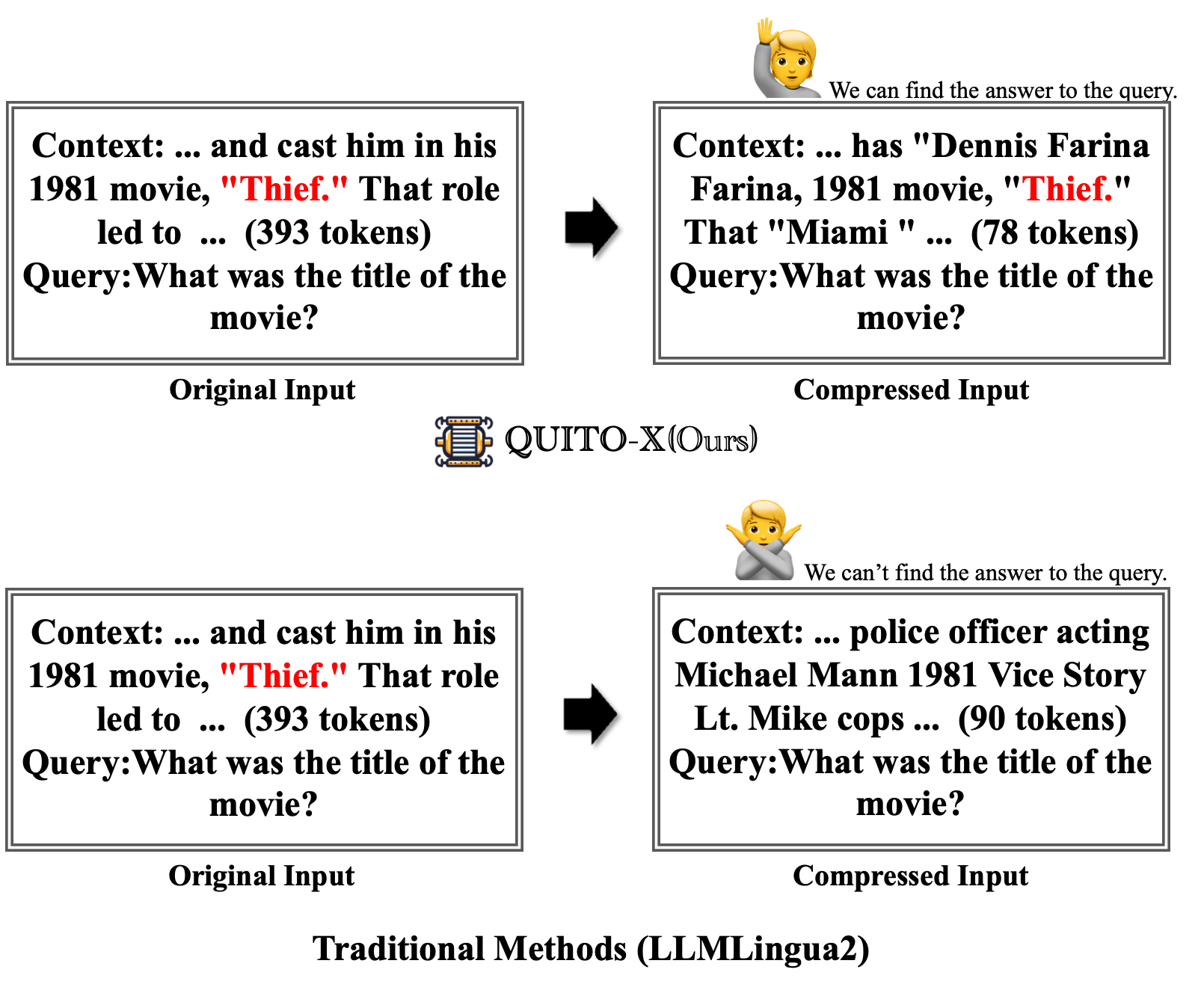}
\centering
\caption{Comparison of our method and baseline approaches for preserving key information in model responses. Our method effectively retains critical context ("Thief"), ensuring accurate interpretation, while baseline methods fail to do so.}
\label{example}
\end{figure}

In recent years, \acp{LLM} \citep{achiam2023gpt} have been widely applied to various tasks in multiple domains, such as text classification \cite{sun2023text}, question answering systems \cite{wang2023self}, and \emph{etc.}. As one of the most promising capabilities of these models, \Ac{ICL} \cite{brown2020language} plays a critical role by enabling the effective use of large language models without requiring additional training. However, in complex tasks, the need to guide the model’s adaptation to the task or provide supplementary knowledge often results in excessively long context, leading to high computational costs, increased inference latency, and the "lost in the middle" problem \cite{tay2020long}. Therefore, how to compress context while maintaining model performance has become a widely studied topic.

\begin{figure}[t!]
\includegraphics[width=0.9\columnwidth]{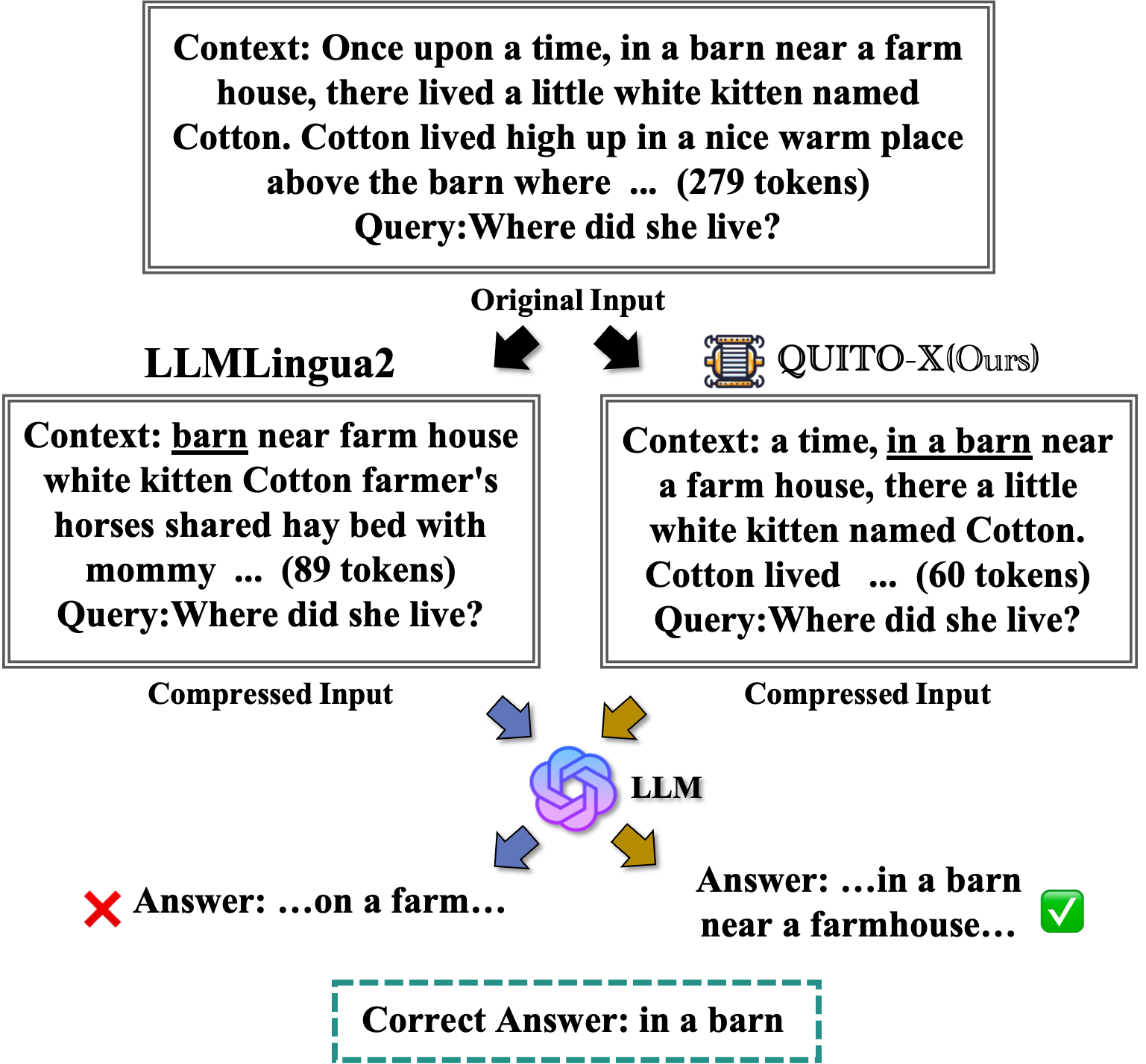}
\centering
\caption{LLMLingua2 overly focuses on high-entropy nouns like 'barn' and 'farmhouse,' while neglecting relational words (e.g., 'near') and verbs, resulting in highly fragmented compression and leading to incorrect answers ('on a farm'). In contrast, QUITO-X retains key relational phrases ('in a barn near a farmhouse'), preserving full meaning and yielding the correct answer.}
\label{case}
\end{figure}

In the literature, \citet{Liu2023RETALLMAR} utilize language models to compress context in a generative manner, while other methods select the most important lexical units (tokens, words, or sentences) from the original context in an extractive manner. Specifically, the generative-based compression methods typically construct compressors by fine-tuning models to generate summaries of the original text, but they are often constrained by inherent limitations of language models, such as restricted context windows, hallucination phenomena, and the "lost in the middle" problem. The extractive-based compression methods is to design appropriate metrics (e.g., self-information \cite{shannon1951prediction}, perplexity (PPL), self-attention) to assign importance scores to each unit, thereby identifying and removing less salient units. However, the metrics used in previous works are not aligned with the optimization goals of the compressor, which may lead to suboptimal results. For example, these metrics often place excessive emphasis on nouns, while overlooking other crucial elements like prepositional phrases, quantifiers or verbs, which may have lower information entropy. However, neglecting such information can result in highly fragmented compression that is difficult to understand, ultimately leading to incorrect model outputs, as shown in Figure \ref{case}.

In this paper, we formulate this problem from an \ac{IB} \cite{tishby2000information, fischer2020conditional} perspective, deriving mutual information as our metric. We also provide a mathematical proof that using mutual information is equivalent to maximizing the likelihood of the compressed output, which is precisely the compressor's optimization objective.
In summary, our contributions are twofold:
\begin{itemize}
    \item \textbf{Applying Information Bottleneck Theory to Context Compression}: We introduce a novel perspective by utilizing Information Bottleneck theory to analyze the properties of context compression. This results in the mutual information metric, and we mathematically prove that it is equivalent to maximizing the likelihood of the compressed generation.
    \item \textbf{Experimental Validation}: We conduct extensive experiments that show significant improvements over previous work on long-context question answering. Moreover, our method reduces memory usage to 50\% of the most memory-efficient baseline while achieving a 25\% improvement in accuracy compared to the best-performing baseline.
\end{itemize}




\section{Related Work}

\subsection{Extractive Context Compression}

Large language models (\acp{LLM}) excel at many tasks but struggle with long inputs due to increased token costs and context truncation. \ac{ICL} \cite{brown2020language} alleviates some of these issues by providing task-relevant prompts but also adds to token usage and inference cost.

To address this, extractive context compression methods remove less relevant tokens or phrases while preserving essential content. Selective Context \cite{li2023compressing} ranks tokens by self-information, while LLMLingua \cite{pan2024llmlingua, jiang2023longllmlingua} compresses input based on \ac{PPL}, using a coarse-to-fine strategy. QUITO \cite{wang2024quito} leverages attention from a small LLM to retain query-relevant context.


These approaches use entropy-based metrics (e.g., self-information, \ac{PPL}) that frequently favor high-entropy tokens such as nouns, while underestimating the importance of function words crucial to relational semantics (Figure~\ref{case}). Furthermore, these metrics are often not theoretically aligned with the underlying optimization objective, such as minimizing KL divergence, thus leading to suboptimal results.

\subsection{Information Bottleneck}

The Information Bottleneck (IB) principle \cite{tishby2000information} aims to compress input $X$ into a representation $T$ that preserves task-relevant information $I(T; Y)$ while discarding irrelevant parts $I(T; X)$:
\begin{equation}
    \mathcal{L}_{\text{IB}} = I(T; X) - \beta I(T; Y).
\end{equation}

In deep learning, IB has been used to interpret representation learning \cite{shwartz2017opening} and inform model compression \cite{alemi2016deep}. In NLP, recent work \cite{zhu2024information} applies IB to filter noisy context for LLMs. Inspired by these works, we build on the information bottleneck principle to derive a token-wise mutual information metric as our optimization objective, using cross-attention scores as a practical proxy. We theoretically prove that this metric is consistent with the maximum likelihood objective, and it achieves state-of-the-art performance across a wide range of long-context evaluation benchmarks.

\section{Method}

\begin{figure*}[t!]
\includegraphics[width=0.8\textwidth]{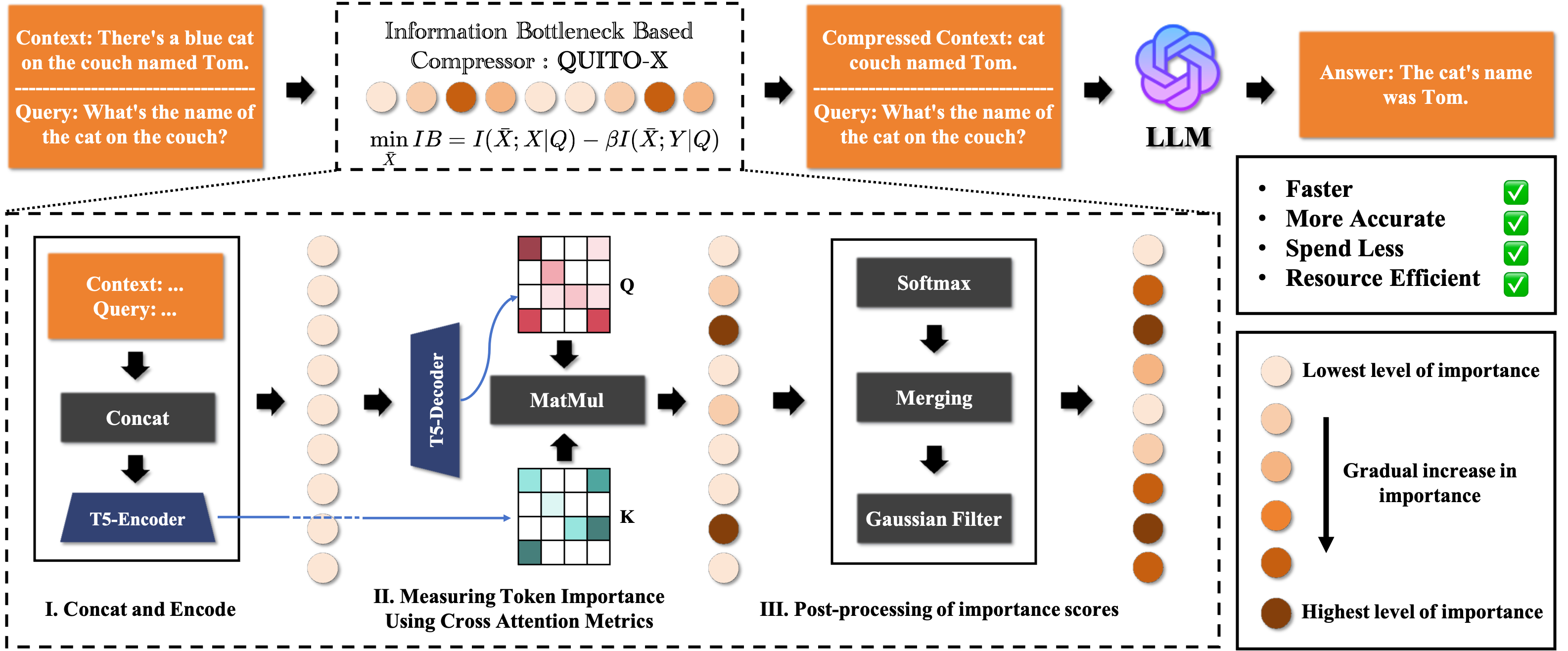}
\centering
\caption{Overview of the proposed method for extracting cross-attention scores using a T5 model. The figure illustrates the process of filtering the context to retain the most relevant information for answering a specific query.}
\label{overview}
\end{figure*}

\subsection{Theorem}
\paragraph{Problem Formulation. }
Given the original context \( X = (x_i)_{i=1}^{L} \) and the query \( Q \), our objective is to filter out unnecessary content from the context \( X = (x_i)_{i=1}^{L} \) into a reduced context \( \bar{X} = (\bar{x}_i)_{i=1}^{\bar{L}} \), while maximizing the likelihood of the ground truth output \( Y \) of the large language model (LLM). This can be formulated as:
\begin{equation}
    \max_{\bar{X}} E \left[ \log \left( P(Y \mid \bar{X}, Q) \right) \right]
\end{equation}
where \( L \) and \( \bar{L} \) represent the sequence lengths of the original context \( X \) and the reduced context \( \bar{X} \), respectively. The compression ratio \( \tau \) is defined as $\tau = \frac{\bar{L}}{L}$
\paragraph{IB Perspective. }
To balance \( \tau \) and the likelihood of \( Y \), we formulate our task as an optimization problem from an information bottleneck perspective\cite{tishby2000information}:
\begin{equation}
    \mathcal{L}_{\text{IB}} = I(\bar{X}; X \mid Q) - \beta I(\bar{X}; Y \mid Q)
\end{equation}
where minimizing the first term improves efficiency, and maximizing the second term ensures correctness.

In the following discussion, we fix the compression ratio \( \tau \) as a constant k. Under this condition, the cost savings from compression are fixed, allowing us to ignore the first term and focus solely on maximizing the second term:
\begin{equation}
    \max_{\bar{X}} I(\bar{X}; Y \mid Q) \quad \text{s.t.} \ \tau = k
\label{eq1}
\end{equation}
The following Theorem 1 demonstrates the consistency between our modeling and the optimization objective of the task.

\paragraph{Theorem 1.}
Under our setting, our optimization objective (\ref{eq2}) is equivalent to (\ref{eq1}):
\begin{equation}
\begin{aligned}
    \max_{\bar{X}} I_Q(\bar{X}; Y) &\sim \max_{\bar{X}} \mathbb{E}[\log P(Y \mid \bar{X}, Q)] \\
    \text{s.t.} \ \tau &= k.
\end{aligned}
\label{eq2}
\end{equation}
The detailed proof is provided in the Appendix \ref{sec:appendixB}.

Using the chain rule of Mutual Information, we have  
\begin{equation}
\begin{aligned}
    I(X;Y \mid Q) &= I_Q(x_1;Y \mid Q) + ... \\ &+ I_Q(x_n;Y\mid x_1,x_2...x_{n-1}, Q)
\end{aligned}
\end{equation}
 Thus, We can break the mutual information between \( X \) and \( Y \) into the mutual information between each token \( x_i \) and \( Y \). we utilize \begin{equation*}
      s(x_i) = I(x_i ; Y \mid x_1,x_2,...x_{i-1},Q) 
 \end{equation*}
 as a metric to measure the importance score of token \( x_i \),  from which we can identify the tokens to retain and those to remove. However, it is difficult to compute the mutual information \( s(x_i) \) directly due to the following reasons:
(i) We cannot access the ground truth output \( Y \) in practical scenarios.
(ii) Even if we use the output of a language model \( Y_\text{LM} \) to approximate \( Y \), the result of \( s(x_i) \) cannot be directly inferred from the probability sampled by the language model.

Therefore, we need to establish a computationally feasible metric to approximate mutual information. Inspired by works in the fields of computer vision and multi-modal learning \cite{dosovitskiy2021imageworth16x16words,esser2024scalingrectifiedflowtransformers}, which often measure the correlation between two types of information \( I_1 \) and \( I_2 \) using either cross-attention between them or self-attention after concatenating \( I_1 \) and \( I_2 \), We conducted several detailed experiments, exploring various strategies for both cross-attention and self-attention, along with other metrics, to determine which method best approximates mutual information. Ultimately, we found that using an encoder-decoder architecture, with \( X \) and \( Q \) as inputs, and leveraging the cross-attention values between the first token of the output \( Y \) and \( x_i \), is the most suitable approach to approximate mutual information in our case. The specific experimental details are provided in the Appendix \ref{sec:appendixA}.

\paragraph{Merging into Lexical Units.}
Following \citet{li2023compressing}, we also merge tokens into words as lexical units to avoid disjoint contexts. We denote \( w \) as a word, \( l_w \) as the length of the word, and \( x_i, x_{i+1}, \dots, x_{i+l_w-1} \) as the tokens comprising the word \( w \) and \( x_{prev} \) represents the preceding context. Benefited from the addition of mutual information, 
\begin{equation}
\begin{aligned}
    &I(x_i,...,x_{i+l_w-1} \mid x_{prev},Y,Q) = I(x_i \mid  x_{prev},Y,Q) \\+& ... + I(x_{i+l_w-1} \mid x_{prev},x_i,...,x_{i+n-2},Q)
\end{aligned}
\end{equation}
we can directly sum the \( s(x_i) \) of all tokens \( x_i \) in a word $w$ to represent \( s(w) \).

\paragraph{Gaussian Smoothing.}
We observed that relying solely on independent metrics for each lexical unit often prioritizes nouns, which typically have high information entropy, while overlooking intermediate conjunctions, verbs, and prepositions. This leads to semantic ambiguity and hampers understanding by large models. To mitigate this issue further, we applied a Gaussian filter on word-level scores
\begin{equation*}
    s(w) = \sum_{k=-K}^K{s(w+k) \cdot g(k)} 
\end{equation*}
\begin{equation*}
    g(k) = \frac{1}{\sigma\sqrt{2\pi}} \exp{(-\frac{k^2}{2\sigma^2})}
\end{equation*}
which helps preserve the information surrounding important units. The detail could be found in section \ref{section 3.2}

\subsection{Algorithm}\label{section 3.2}

Our method compresses long contexts into concise, informative representations through three key steps, as shown in Figure \ref{overview}:

\textbf{Concat and Encode:}  
    The \( X \) and \( Q \) are concatenated into a single input sequence \( X+Q \) and fed into the \( f_{enc} \). This produces a sequence of hidden representations that captures the semantic and positional information of the input tokens:
    \begin{equation}
        \{h_t\} = f_{enc}(X+Q)
    \end{equation}
    Here, \( h_t \) represents the hidden representation of the \( t \)-th token.

\textbf{Measuring Token Importance:}  
    During the decoding process, the cross-attention mechanism \( f_{attn} \) is leveraged to compute the importance of each token in the context relative to the query. Specifically, hidden representation of the decoder's first token \( h_{<start>} \) attends to all tokens in the encoded sequence via the cross-attention mechanism:
    \begin{equation}
        \{a_t\} = f_{attn}(\{h_t\}, h_{<start>})
    \end{equation}
    Here, \( a_t \) denotes the attention score assigned to the \( t \)-th token, reflecting its relative importance with respect to the query.

\textbf{Post-processing of Importance Score:}  
    The attention weights for context tokens are extracted, averaged across all attention heads, and normalized using a softmax function. 
    \begin{equation}
        s(t) = \frac{\exp{a_t}}{\sum_{token\in f_{tok}(X)}{\exp{a_{token}}}}, t \in f_{tok}(X)
    \end{equation}
    We use \( f_{tok} \) for tokenization, these scores represent the relevance of each token in the tokenized context to the given query.
    
    The normalized token scores are aggregated at the word level:  
    \begin{equation}
        s(w) = \sum_{t \in w}{s(t)}, w \in X
    \end{equation}
    To account for the contextual importance of words, a Gaussian filter is applied to the word-level scores. This ensures that words appearing near important terms also receive elevated scores:
    \begin{equation}
    s(w) = \sum_{k=-K}^K{s(w+k) \cdot g(k)}
    \end{equation}
    \begin{equation}
    g(k) = \frac{1}{\sigma\sqrt{2\pi}} \exp{(-\frac{k^2}{2\sigma^2})}
    \label{eq3}
    \end{equation}
    
    Based on the smoothed scores, we retain only the most relevant words to form the compressed context. The compression ratio \( \tau \) can be adjusted to control the level of detail retained. The function \( f_{top} \) selects words whose scores are among the top \( \tau \) proportion:
    \begin{equation}
        \bar{X} = f_{top}(\{s(w)\}, \tau), w \in X
    \end{equation}

This algorithm effectively reduces context length while retaining essential information, ensuring accurate and efficient performance in downstream tasks.

\input{figures/compare.tex}

\section{Experiments}

\subsection{Datasets and Metrics}

We conduct experiments on nine datasets that vary in text length and task type, covering both manageable and excessively long contexts:

(i) \textbf{CoQA \cite{reddy2019coqa} and Quoref \cite{dasigi2019quoref}:} These datasets feature texts of moderate length, within the processing capability of large models, making them ideal for standard evaluations of model performance.

(ii) \textbf{2WikiMultiHopQA \cite{ho2020constructing}, HotpotQA \cite{yang2018hotpotqa}, MuSiQue \cite{trivedi2022musique}, TriviaQA \cite{joshi-etal-2017-triviaqa}, and Gov\_Report \cite{huang-etal-2021-efficient}:} These datasets are part of the LongBench benchmark \cite{bai2023longbench}, which focuses on long-context understanding across diverse NLP tasks such as multi-doc QA, few-shot QA, and summarization. They typically feature excessively long inputs that challenge models' ability to retain and reason over relevant information, often suffering from the "lost in the middle" phenomenon.

To evaluate model accuracy, we adopt the \ac{EM} metric for question answering datasets, which measures the percentage of predictions that exactly match the ground truth answers. For the summarization dataset Gov\_Report, we report ROUGE-L \cite{lin2004rouge}, a widely used metric that assesses the overlap between generated summaries and reference summaries.

\input{figures/main1.tex}

\input{figures/main3.tex}

\subsection{Implementation Details}

We employed the FLAN-T5-small model \cite{chung2024scaling} for compression. Our approach leverages Huggingface Transformers and PyTorch 2.1.0 with CUDA-12.1. For question-answering tasks, we utilized LongChat-13B-16k \cite{li2023long} and LLaMA3-8B-Instruct \cite{llama3modelcard}.

In our experiments, we observed that the choice of the parameter \(\sigma\) in (\ref{eq3}) does not significantly impact the compression performance as long as \(\sigma \neq 0\). Therefore, for consistency, we set \(\sigma = 1\) for all subsequent experiments. Detailed parameter search results are provided in the Appendix \ref{sec:appendixD}.

For CoQA \cite{reddy2019coqa} and Quoref \cite{dasigi2019quoref}, we evaluated model accuracy using the original context and without any context, aiming to assess the models' ability to summarize with full information and rely on prior knowledge. Next, we tested five baseline methods and our proposed approach at compression ratios of 0.75, 0.50, and 0.25, measuring accuracy with the compressed context using both LongChat-13B-16k and LLaMA3-8B-Instruct models.  

For datasets with long contexts, including 2WikiMultiHopQA \cite{ho2020constructing}, HotpotQA \cite{yang2018hotpotqa}, and MuSiQue \cite{trivedi2022musique}, TriviaQA \cite{joshi-etal-2017-triviaqa}, Gov\_Report \cite{huang-etal-2021-efficient}, sourced from LongBench \cite{bai2023longbench}, we focused on the LLaMA3-8B-Instruct model. To handle the extreme length of these texts, a chunking strategy was adopted, dividing the context into 512-token chunks. Two strategies were tested:

\noindent\textbf{Strategy 1:} Compressing each chunk individually and then merging the compressed representations.

\noindent\textbf{Strategy 2:} Calculating attention scores between each chunk and the query, merging these attention scores across all chunks, and then performing a unified compression on the merged context.
\subsection{Baseline}
We compared against the following context compression baselines in Table \ref{tab:compare}: (1) \textbf{Selective Context} \cite{li2023compressing}: Uses GPT-2 \cite{radford2019language} to retain context segments based on self-information. (2) \textbf{LLMLingua} \cite{pan2024llmlingua}: Employs Llama-2-7b \cite{touvron2023llama} with dynamic compression driven by context \ac{PPL}. (3) \textbf{LongLLMLingua} \cite{jiang2023longllmlingua}: Extends LLMLingua for longer contexts, also using Llama-2-7b \cite{touvron2023llama}. (4) \textbf{LLMLingua2} \cite{pan2024llmlingua}: Utilizes XLM-RoBERTa-large \cite{conneau2019unsupervised}, introducing data distillation for compression. (5) \textbf{QUITO} \cite{wang2024quito}: Applies Qwen2-0.5B-Instruct \cite{yang2024qwen2} with attention mechanisms to selectively retain query-relevant context.

For datasets with manageable text lengths, such as \textbf{CoQA} \cite{reddy2019coqa} and \textbf{Quoref} \cite{dasigi2019quoref}, we evaluated our method against all listed baselines. These datasets allowed us to test the effectiveness of each approach in compressing contexts without encountering extreme text length challenges.  

For datasets with long contexts, including \textbf{2WikiMultiHopQA} \cite{ho2020constructing}, \textbf{HotpotQA} \cite{yang2018hotpotqa}, \textbf{MuSiQue} \cite{trivedi2022musique}, \textbf{TriviaQA} \cite{joshi-etal-2017-triviaqa}, and \textbf{Gov\_Report} \cite{huang-etal-2021-efficient}, we focus our comparison on \textbf{LLMLingua2}, as well as two additional baselines: \textbf{Selective Context} and \textbf{Quito}. These datasets pose different challenges: the QA datasets (multi-doc QA and few-shot QA) often suffer from the ``lost in the middle'' phenomenon, while the summarization dataset (\textbf{Gov\_Report}) requires models to preserve critical information across lengthy documents. Together, they provide a comprehensive evaluation of our method's performance in long-context scenarios across diverse task types.

\begin{figure*}[bt]
\includegraphics[width=1\textwidth]{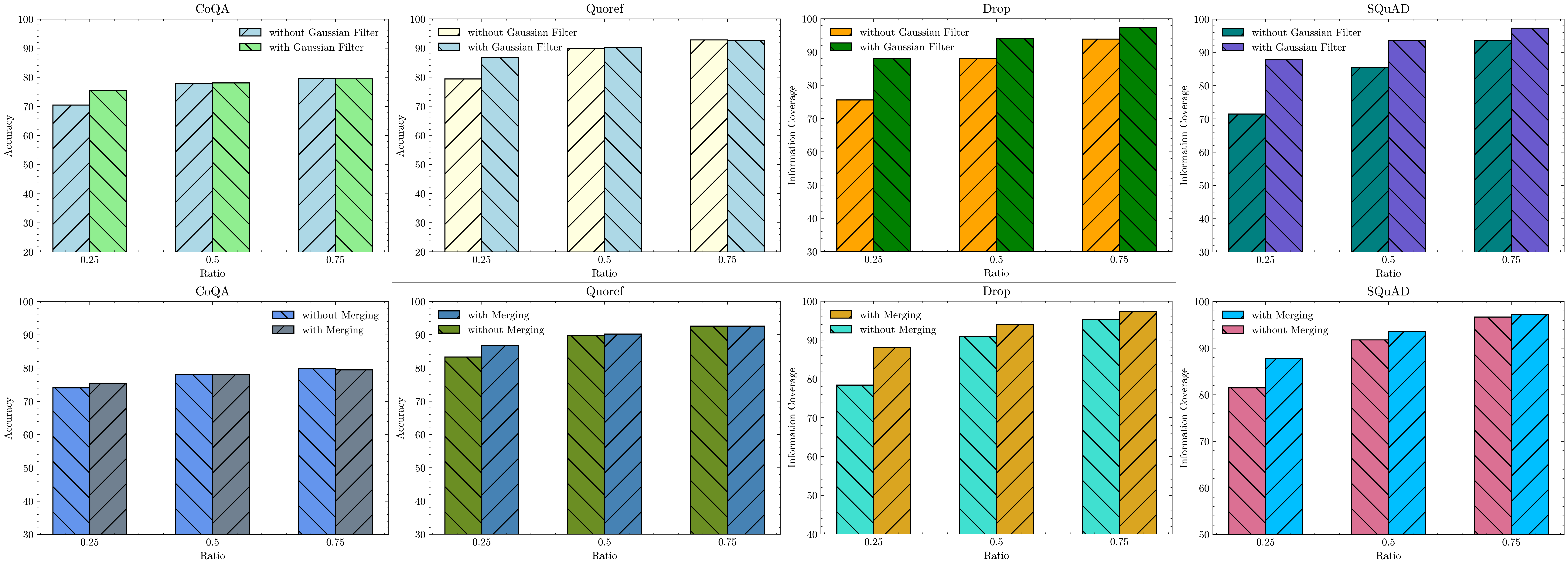}
\centering
\caption{Ablation study results on four datasets (CoQA, Quoref, DROP, SQuAD) under three compression ratios (0.25, 0.5, 0.75). The top row shows the impact of the Gaussian filter on accuracy and information coverage, demonstrating consistent improvements across all datasets and compression ratios. The bottom row illustrates the effect of the merging module, highlighting its importance in recovering meaningful representations, particularly under higher compression ratios.}
\label{ablation}
\end{figure*}

\subsection{Experimental Results}

The results shown in Table \ref{tab:main-result} and Table \ref{long} comprehensively demonstrate the effectiveness of our proposed methods across various datasets and compression ratios.

For the Quoref and CoQA datasets (Table \ref{tab:main-result}), our proposed \textbf{\our} consistently outperforms existing baselines, including Selective-Context, LLMLingua, LongLLMLingua, LLMLingua2, and QUITO, under all tested compression ratios (1.00, 0.75, 0.50, 0.25, and 0.00). Remarkably, \textbf{\our} achieves superior performance even at higher compression ratios, where significant portions of context are removed. This robust performance highlights the capability of our method in retaining critical information despite substantial context reductions. In some cases, particularly noted in the underlined sections of Table \ref{tab:main-result}, our method even surpasses the performance of the original, uncompressed context. This suggests that our approach not only removes irrelevant noise but also enables the model to focus better on relevant portions of the context, thereby improving prediction quality.

For long-text datasets (Table \ref{long}), including 2WikiMultiHopQA, HotpotQA, MuSiQue, TriviaQA, and Gov\_Report, the supplementary experiments further validate the adaptability and robustness of our strategies under varying compression levels.

In the multi-doc QA datasets (2WikiMultiHopQA, HotpotQA, and MuSiQue), both proposed strategies (\textbf{Strategy 1} and \textbf{Strategy 2}) consistently outperform the baselines. For example, in 2WikiMultiHopQA, \textbf{Strategy 1} achieves the best result at a compression ratio of 0.75, while \textbf{Strategy 2} excels at 0.50. In HotpotQA, \textbf{Strategy 2} demonstrates the highest performance at 0.25 and 0.50 ratios. In MuSiQue, \textbf{Strategy 2} shows a clear advantage at lower ratios, particularly under the most aggressive compression (0.25).

On the few-shot QA dataset TriviaQA, our method also achieves consistent improvements over baselines across different compression ratios. This result highlights the effectiveness of our approach even in scenarios with limited supervision and long input contexts.

For the summarization dataset Gov\_Report, our method yields higher ROUGE-L scores compared to other baselines, particularly under medium and high compression levels. This demonstrates that our strategy not only maintains key information but also preserves summary quality even with significantly reduced context, which is especially important in summarization tasks involving lengthy documents.

These results collectively underscore the robustness, adaptability, and overall effectiveness of our proposed methods for handling compressed contexts across a variety of datasets, task types, and compression scenarios.

\subsection{Ablation Study}

\paragraph{Gaussian Filter.}  
The top row of Figure \ref{ablation} shows the effect of the Gaussian filter across different datasets and compression ratios (0.25, 0.5, 0.75). For CoQA and Quoref, we use accuracy as the evaluation metric, while for DROP and SQuAD, we adopt information coverage, which we explain further in the Appendix \ref{sec:appendixC}. The Gaussian filter consistently improves performance, particularly at lower ratios. For example, in SQuAD, information coverage increases significantly (from 71.5 to 87.8) at the 0.25 ratio. These results demonstrate its effectiveness in retaining critical context information during compression.

\paragraph{Merging.}  
The bottom row of Figure \ref{ablation} highlights the impact of the merging module. Merging consistently boosts accuracy and information coverage, especially at the 0.25 ratio where context loss is severe. For instance, in DROP, merging improves information coverage by nearly 10 points. This confirms its role in preserving meaningful context under high compression.

\subsection{Comparison with Sentence-Level Compression}

To further evaluate the effectiveness of our token-level compression approach, we compare it against FILCO \cite{wang2023learning}, a sentence-level method that compresses long contexts by selecting salient sentences. We follow FILCO's experimental protocol and preprocessing pipeline on NQ and TQA, using their released datasets and settings to ensure a fair comparison.

As shown in Appendix~\ref{appendix:appendixG}, our method outperforms FILCO under comparable compression ratios (25\% and 50\%) on both datasets.

\section{Conclusion}

In this paper, we aim to tackle the challenge of context compression. Leveraging information bottleneck theory, we derive mutual information as the optimization objective, which we prove to be equivalent to maximizing likelihood. Our method significantly outperforms strong baselines in both inference latency and performance. Furthermore, it excels on long texts, occasionally surpassing models that utilize the original context, likely by eliminating inherent redundancy in the context. More effective chunking strategies for long texts are left for future exploration.

\section*{Limitations}

Despite the strong performance and efficiency gains demonstrated by our method, there are several limitations worth noting:

First, due to the restricted context window of smaller language models, our approach relies on chunking strategies to process long documents. While this proves effective across many datasets, chunking inevitably breaks the global context and may lead to semantic discontinuities between chunks. How to maintain coherence across chunk boundaries—or to quantify the impact of such fragmentation—remains an open research question.

Second, since our method performs compression at the token level, the resulting outputs can suffer from reduced human readability. Compared to sentence-level or summarization-based methods, token-level outputs tend to appear fragmented or syntactically incomplete. Although this does not impair the model’s ability to interpret the compressed input and answer questions accurately, it may reduce the interpretability of the compression decisions from a human perspective. 

That said, as we demonstrate in Appendix~\ref{appendix:appendixH}, our approach retains significantly better semantic continuity and readability compared to other token-level baselines (e.g., LLMLingua2 and QUITO). This highlights the potential of our method to strike a balance between compression granularity and human interpretability. Future work may explore ways to further enhance this trade-off, for example by integrating syntactic structure or discourse markers into the token selection process.

Finally, due to computational constraints, we were unable to conduct broader-scale experiments across more diverse domains. As a result, certain hyperparameters—such as the Gaussian smoothing parameter $\sigma$—have not been comprehensively tuned across all datasets. While our experiments suggest the method is relatively stable under reasonable variations of $\sigma$, further large-scale validation would strengthen the generalizability claims.

\section*{Acknowledgments}
The authors wish to thank the anonymous reviewers for their helpful comments. This work was supported by the Natural Science Foundation of China (key program) [grant number 62441229], the Beijing Natural Science Foundation [grant number 4252022], and the National Key R\&D Program of China [grant number 2022YFB2404200].

\bibliography{custom}

\appendix

\section{Experimental Selection of Mutual Information Metric}

\label{sec:appendixA}
\subsection{Motivating Observation}
To identify a metric that best approximates the mutual information $I(X; Y \mid Q)$, we designed the following experiment: we filtered a subset from the Drop QA dataset, denoted as $\mathcal{D} = \{\mathcal{D}_i\}_{i=1}^n=\{X_i, Y_i, Q_i\}_{i=1}^n$. In $\mathcal{D}$, $Y_i$ is a substring of $X_i$. The substring $Y_i$ within $X_i$ (hereafter referred to as $\text{Sub}_{Y_i}$) captures the majority of the mutual information between $X_i$ and $Y_i$. Informally, the higher the relative value of a metric on the tokens of these substrings, the better the metric can measure $I(X; Y \mid Q)$.

\subsection{Experiment}
We tested several commonly used metrics, including self-attention \cite{wang2024quito} and self-information \cite{li2023compressing}. Cross-attention is a prevalent metric for measuring the correlation between two pieces of information. We used Flan-T5-small \cite{chung2024scaling} to compute cross-attention and implemented the following two strategies for each $\mathcal{D}_i$:
\paragraph{cross attn first.}Compute only the cross-attention scores between the first token \texttt{\textless start\textgreater} in $Y_i$ and each token in $X_i$.
\paragraph{cross attn total.}Autoregressively generate $Y_i$ and compute the average sum of the cross-attention scores between all tokens in $Y_i$ and all tokens in $X_i$.

We adopted Mean Reciprocal Rank (MRR) \cite{10.1145/371920.371973,radev-etal-2002-evaluating} to evaluate which metric better represents mutual information. Specifically, for each metric, we first calculate the MRR for each data point $\mathcal{D}_i = \{X_i, Y_i, Q_i\}$ individually. For a given $\mathcal{D}_i$, we calculate the value of each token based on the metric, sort them to obtain their rank array, and then compute MRR assuming $\text{Sub}_{Y_i}$ has a length of $len$ and appears at positions $k, \ldots, k+len-1$:
\[
\text{MRR}_i = \frac{1}{len} \sum_{j=1}^{len} \frac{1}{\text{rank}_{k+j-1}}
\]
Finally, the overall MRR for the dataset $\mathcal{D}$ is obtained by averaging $\text{MRR}_i$ across all data points:
\[
\text{MRR} = \frac{1}{|\mathcal{D}|} \sum_{i=1}^{|\mathcal{D}|} \text{MRR}_i
\]

\subsection{Result}
\begin{figure}[t!]
\includegraphics[width=\columnwidth]{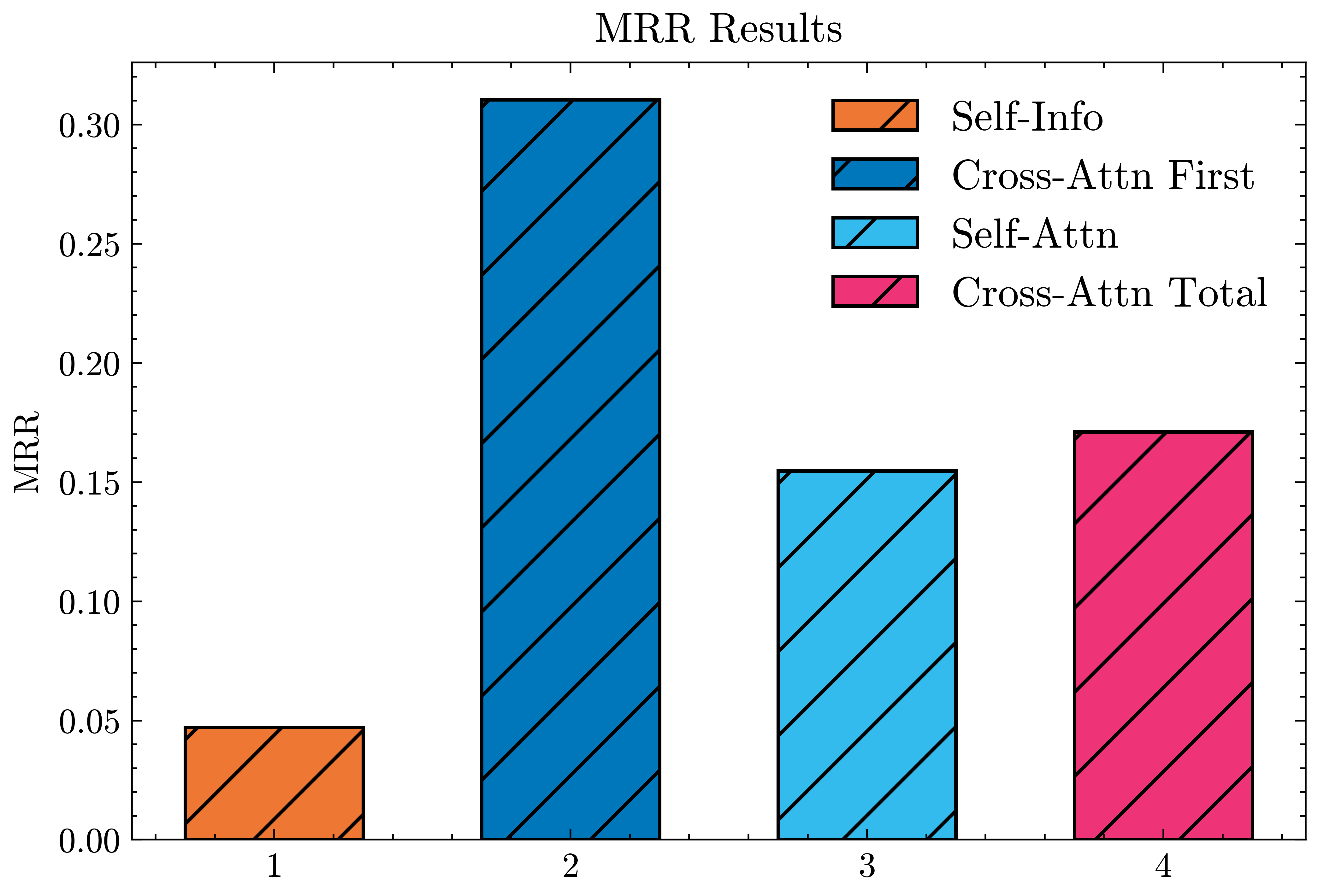}
\caption{MRR results}
\label{mrr}
\end{figure}

The experimental results are presented in Figure ~\ref{mrr}. The results indicate that using the cross-attention value between the first token of output $Y$ and each $x_i$ yields a significantly higher MRR compared to other methods.

\section{Proof of Theorem 1}
\label{sec:appendixB}
Let \( X \) be the original context, \( Q \) be the query, \( Y \) be the output, and \( \bar{X} \) be the extractive compressed result. Denote \( \tau \) as the compression rate, and let \( k \) be a constant such that \( k \in (0, 1] \).
\paragraph{Theorem}
\begin{equation}
\begin{aligned}
    \max_{\bar{X}} I_Q(\bar{X}; Y) &\sim \max_{\bar{X}} \mathbb{E}[\log P(Y \mid \bar{X}, Q)] \\
    \text{s.t.} \ \tau &= k.
\end{aligned}
\end{equation}
(To simplify the notation, we use \( I_Q \) to represent the condition on \( Q \).)

\paragraph{Proof:}
We start by expanding the mutual information term \( I_Q(\bar{X}; Y) \):
\begin{align*}
&I_Q(\bar{X}; Y) = \\ &\int_{\bar{x}, y, q} P(\bar{x}, y \mid q) \log \left( \frac{P(\bar{x}, y \mid q)}{P(\bar{x} \mid q) P(y \mid q)} \right) d\bar{x} \, dy \, dq \\
&= \int_{\bar{x}, y, q} P(\bar{x}, y \mid q) \log \left( \frac{P(\bar{x}, y \mid q)}{P(\bar{x} \mid q)} \right) d\bar{x} \, dy \, dq \\&- \int_{y, q} \log P(y \mid q) (\int_{\bar{x}}P(\bar{x},y \mid q) d\bar{x})\, dy \, dq \\
&= \int_{\bar{x}, y, q} P(\bar{x}, y \mid q) \log \left( \frac{P(\bar{x}, y \mid q)}{P(\bar{x} \mid q)} \right) d\bar{x} \, dy \, dq \\&- \int_{y, q} \log P(y \mid q) P(y \mid q) \, dy \, dq
\end{align*}
Since $\int_{y, q} \log P(y \mid q) P(y \mid q) \, dy \, dq$ does not affect the optimization, we ignore it:
\begin{align*}
&I_Q(\bar{X}; Y) 
\\&\sim \int_{\bar{x}, y, q} P(\bar{x}, y \mid q) \log \left( \frac{P(\bar{x}, y \mid q)}{P(\bar{x} \mid q)} \right) d\bar{x} \, dy \, dq \\
&= E_{\bar{X}, Y, Q} \left[ \log P(y \mid \bar{x}, q) \right].
\end{align*}
Here \(\bar{x}, y, q\) represent specific data points sampled from the random variables \(\bar{X}, Y, Q\), respectively.  This completes the proof.

\section{Information Coverage}
\label{sec:appendixC}

In this section, we explain the Information Coverage metric used in our ablation study for DROP and SQuAD datasets. Unlike accuracy, which directly measures the correctness of the model's predictions, Information Coverage focuses on whether key information (i.e., the source of the answer) is preserved after context compression.

\begin{figure}[t!]
\includegraphics[width=\columnwidth]{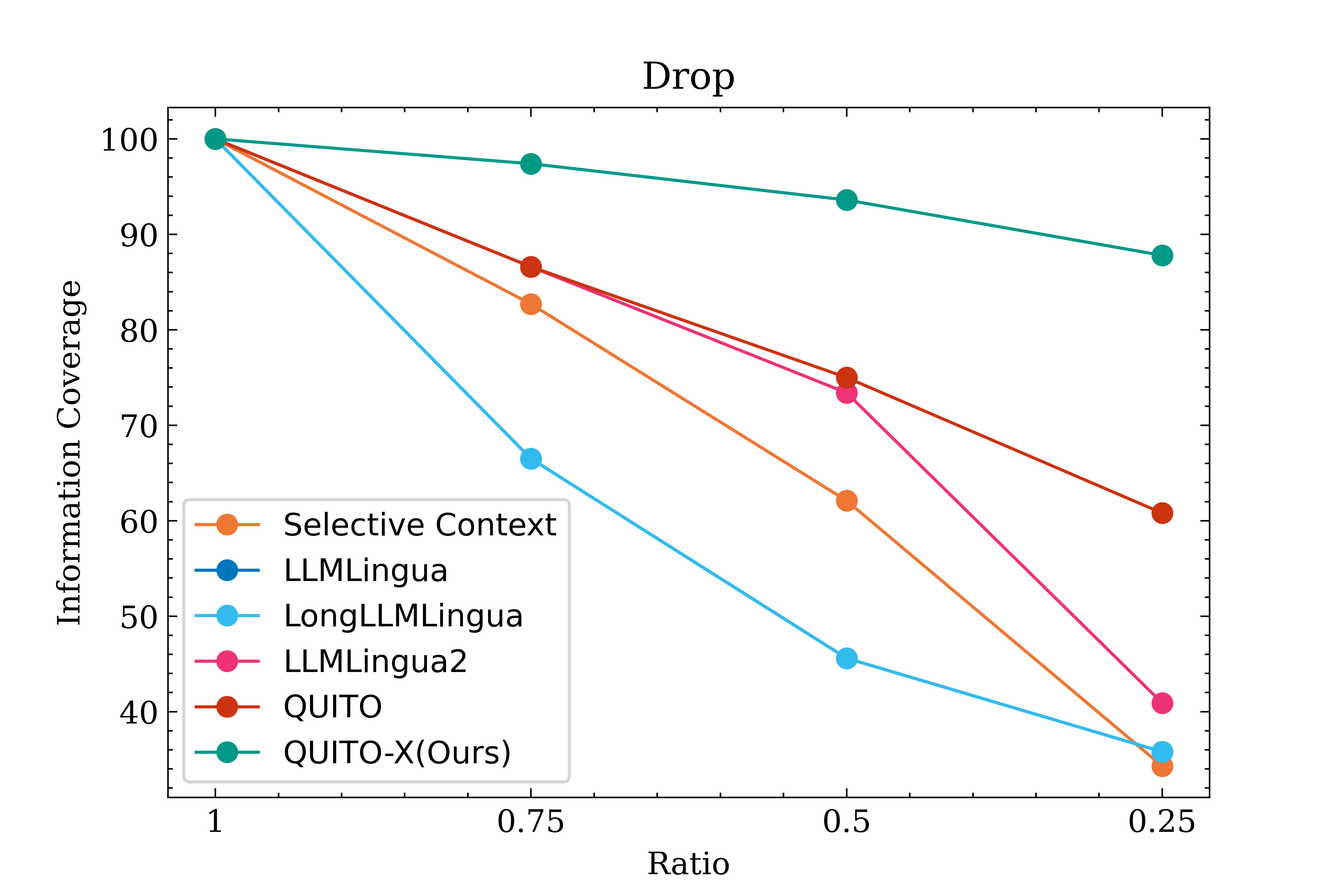}
\caption{Information coverage on Drop.}
\label{drop_ic}
\end{figure}

\begin{figure}[t!]
\includegraphics[width=\columnwidth]{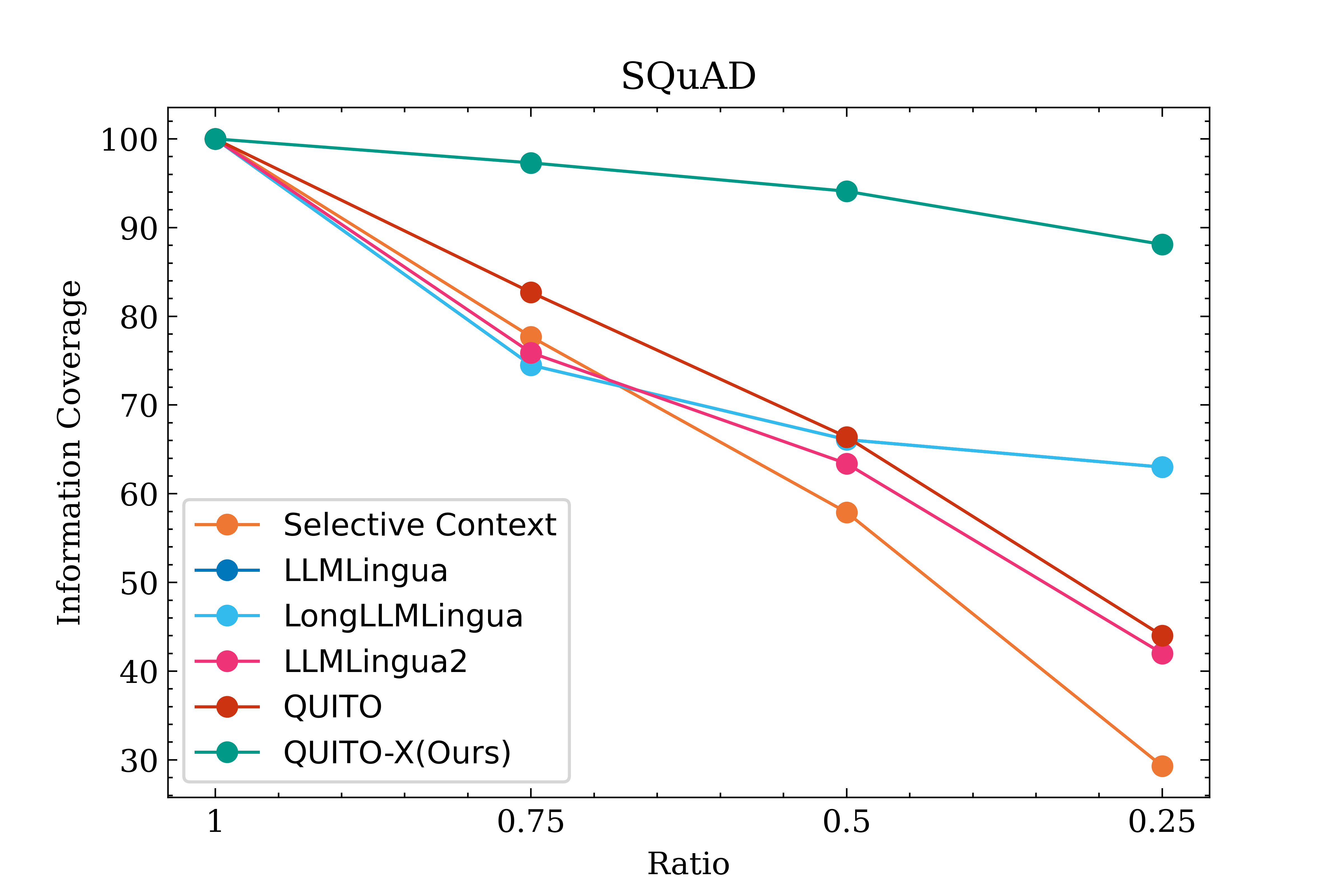}
\caption{Information coverage on SQuAD.}
\label{squad_ic}
\end{figure}

Specifically, we adopt \ac{EM} as the evaluation metric for measuring coverage. Given a compressed context and a target answer, EM evaluates whether the answer's source can still be precisely matched within the compressed context. This ensures that critical information needed to derive the answer is retained post-compression. A higher EM score indicates better preservation of essential information, thus reflecting the compression method's effectiveness in maintaining important content.

Figures \ref{drop_ic} and \ref{squad_ic} showcase the Information Coverage at different compression ratios (from 1.0 to 0.25) on the DROP and SQuAD datasets. These results are independent of the ablation experiments and are intended to highlight the robustness of our proposed method under varying levels of compression.

From the figures, it is evident that across all compression ratios, our method consistently achieves the highest Information Coverage compared to baseline approaches. This demonstrates the effectiveness of our method in preserving critical answer-related information, even as the context length is reduced. Notably, at lower compression ratios (e.g., 0.25), where information loss is more severe, our approach still outperforms other methods by a clear margin, underscoring its ability to prioritize and retain essential content.

These findings further confirm that our method can effectively mitigate the challenges of information loss during compression while maintaining performance in downstream tasks.

\section{Parameter Search for \(\sigma\)}
\label{sec:appendixD}

In our experiments, we examined the effect of different values of the parameter \(\sigma\) on the performance of the compression technique. Specifically, \(\sigma\) controls the variance of the Gaussian filter used during context compression. To explore its impact, we conducted a parameter search across several values of \(\sigma\), ranging from 1 to 5, to assess how variations in \(\sigma\) influence model performance at different compression ratios.

\begin{figure*}[t!]
\includegraphics[width=\textwidth]{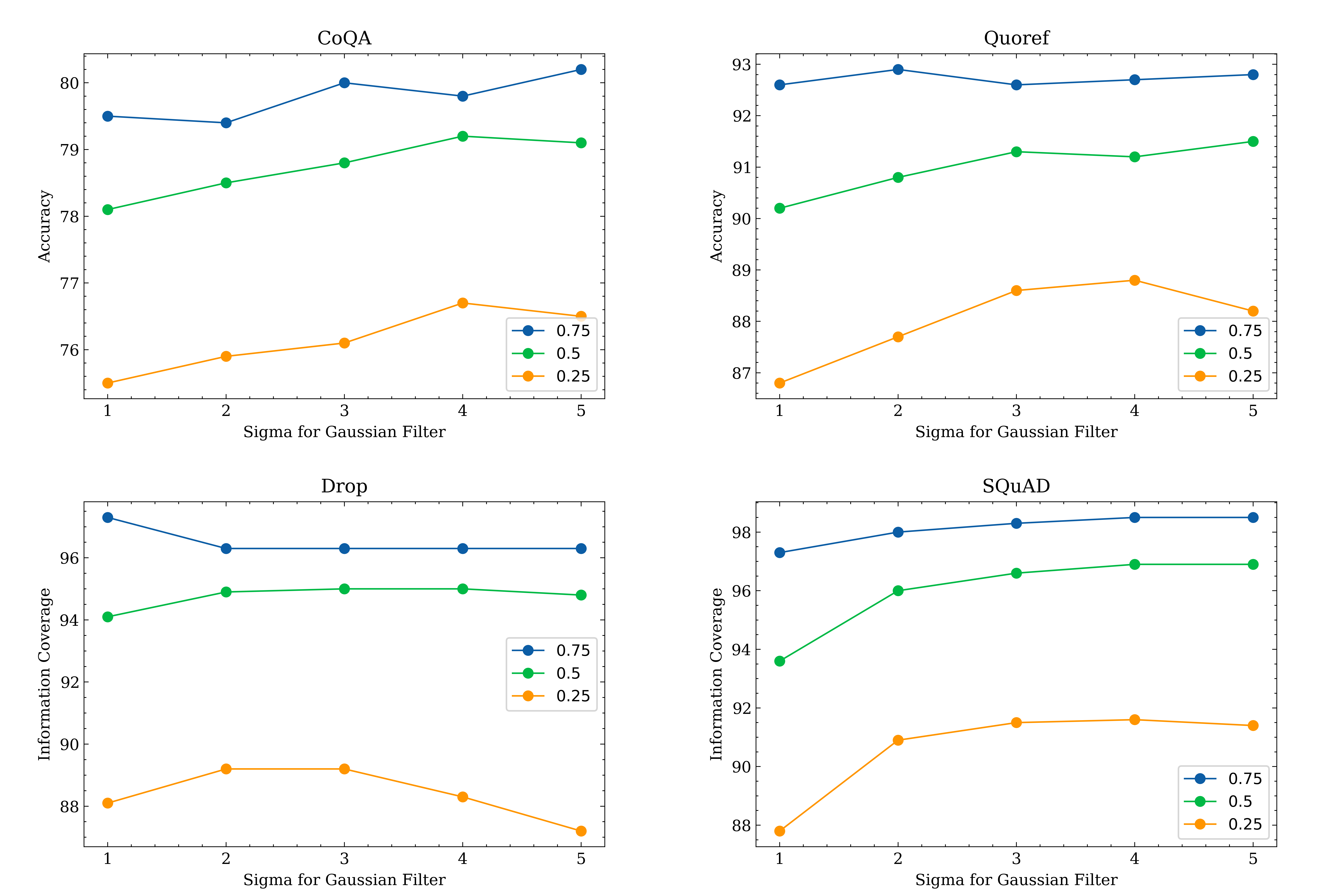}
\caption{parameter search across several values of \(\sigma\)}
\label{param}
\end{figure*}

Figure \ref{param} shows the results of this search, where we measured the model's accuracy and information coverage at compression ratios of 0.75, 0.50, and 0.25. 

From our observations, we found that the value of \(\sigma\) had minimal impact on performance for non-zero values, with only a slight variation in both accuracy and information coverage. Based on these findings, we chose \(\sigma = 1\) as the default value for all subsequent experiments, ensuring both consistent and efficient compression without substantial loss in performance.

For a detailed breakdown of the parameter search, see the plot in Figure \ref{param}, which illustrates how \(\sigma\) affects model performance across all datasets tested.

\section{Computational Overhead Analysis}
\label{sec:appendixE}

The computational overhead of our approach primarily arises from calculating the cross-attention during inference with a relatively small proxy model. Similarly, the PPL-based method incurs additional time overhead from computing log-likelihood during inference using the same proxy model. In both methods, the time overhead is approximately equivalent to one round of inference by the proxy model.

\subsection{Inference Time per 512 Tokens}

The table below details the inference time per 512 tokens for different models:

\begin{table}[h!]
\centering
\begin{tabular}{|c|c|}
\hline
\textbf{Model} & \textbf{Time per 512 Tokens} \\
\hline
Llama3-8B     & 2.4251s \\
Flan-T5-Small & 0.3238s \\
\hline
\end{tabular}
\caption{Inference time per 512 tokens for different models.}
\end{table}

For our method, we use FLAN-T5-Small, a model with only 80M parameters, as the proxy model. This makes the additional time overhead negligible. The efficiency gains from our approach far outweigh this minimal time cost. Furthermore, it is important to note that while our method and the PPL-based method theoretically share the same additional time cost when employing the same proxy model, prior works typically use much larger models as proxies. This makes our method more efficient in practice.

\section{Comparison with Different FLAN-T5 Model Sizes}

To demonstrate the versatility of our approach, we compared models with different sizes of the encoder-decoder architecture. Specifically, we used various models from the Flan-T5 series (Flan-T5-small, Flan-T5-base, Flan-T5-large), as there are no other encoder-decoder models that rival Flan-T5 within the same time frame. Older models like BART (2019) and T5 (2019) show a significant performance gap compared to Flan-T5. For efficiency reasons, we primarily utilized Flan-T5-Small in our experiments. We also benchmarked Flan-T5-Base and Flan-T5-Large, with their results showing similarly promising trends, as shown in the table \ref{tab:5}.

\begin{table}[h!]
\centering
\begin{tabular}{|c|c|c|c|c|}
\hline
\textbf{Ratio} & \textbf{Dataset} & \textbf{Small} & \textbf{Base} & \textbf{Large} \\
\hline
0.75  & Squad    & 97.3   & 98.3   & 98.2   \\
0.5   &          & 94.1   & 96.4   & 95.6   \\
0.25  &          & 88.1   & 92.1   & 90.4   \\
\hline
0.75  & Quoref   & 92.6   & 92.4   & 92.2   \\
0.5   &          & 90.2   & 90.1   & 90.3   \\
0.25  &          & 86.8   & 89.4   & 89.9   \\
\hline
0.75  & CoQA     & 79.5   & 80.3   & 80.1   \\
0.5   &          & 78.1   & 78.6   & 79.9   \\
0.25  &          & 75.5   & 77.8   & 77.5   \\
\hline
\end{tabular}
\caption{Evaluation results for different sizes of FLAN-T5 models on various datasets.}
\label{tab:5}
\end{table}

\section{Comparison with Sentence-Level Compression Methods}
\label{appendix:appendixG}

To compare our token-level compression with sentence-level methods, we replicate FILCO’s experimental setup on NQ and TQA, two question answering benchmarks with long input contexts. We use the same preprocessed datasets and evaluation protocol as described in FILCO’s original paper to ensure fair comparison.

Table~\ref{tab:filco} summarizes the results under 25\% and 50\% compression ratios.

\begin{table}[ht]
\centering
\begin{tabular}{lcc}
\toprule
\textbf{Method} & \textbf{NQ} & \textbf{TQA} \\
\midrule
FILCO (44\%-64\%) & 44.24 & 59.50 \\
Ours (50\%) & \textbf{60.91} & 60.19 \\
Ours (25\%) & 56.79 & \textbf{60.95} \\
\bottomrule
\end{tabular}
\caption{Comparison with sentence-level compression (FILCO) on NQ and TQA under 25\% and 50\% compression. Our method consistently outperforms FILCO.}
\label{tab:filco}
\end{table}

Compared to sentence-level approaches like FILCO, our method achieves superior performance and offers precise compression rate control, making it particularly effective in low-budget scenarios where retaining critical information is crucial.

\section{Case Studies on Readability and Semantic Continuity}
\label{appendix:appendixH}

To evaluate the readability and semantic integrity of the compressed outputs, we conducted case studies comparing our method with several strong baselines, including LLMLingua2, QUITO, and their variants. Figure~\ref{fig:casestudy1} and \ref{fig:casestudy2} illustrate representative examples.

\begin{figure*}[htb]
\begin{tcolorbox}
\textbf{Original Prompt (Census QA Example)}:\\
As of the census of 2000, there were 218,590 people, 79,667 households, and 60,387 families residing in the county.  The population density was 496 people per square mile (192/km²). There were 83,146 housing units at an average density of 189 per square mile (73/km²). The racial makeup of the county was 86.77\% Race (United States Census), 9.27\% Race (United States Census), 0.23\% Race (United States Census), 1.52\% Race (United States Census), 0.06\% Race (United States Census), 0.69\% from Race (United States Census), and 1.47\% from two or more races.  1.91\% of the population were Race (United States Census) or Race (United States Census) of any race. 22.5\% were of German people, 13.1\% Irish people, 9.8\% Italian people, 9.2\% English, 8.1\% "American" and 6.0\% Polish ancestry.\\
\textbf{Question:} Which group from the census is smaller: German or English?
\end{tcolorbox}

\begin{tcolorbox}
\textbf{Compressed Prompt (LLMLingua2)}:\\
2000, 218,590 79,667 households 60,387 families 496 83,146 units 189 racial makeup 86.77\% 1.47\% 1.91\% 22.5\% German 13.1\% 9.8\% Italian 9.2\% 8.1\% 6.0\% Polish
\end{tcolorbox}

\begin{tcolorbox}
\textbf{Compressed Prompt (QUITO)}:\\
2000, 79,667 households, and 60,387 families residing There were 86.77\% Race (United Race race. 22.5\% of German people, 13.1\% Irish people, 6.0\% Polish ancestry.
\end{tcolorbox}

\begin{tcolorbox}
\textbf{Compressed Prompt (Ours)}:\\
the people, 79,667 households, and 60,387 families residing 22.5\% of German people, 13.1\% Irish people, 9.8\% Italian people, 9.2\% English, 8.1\% "American" and 6.0\% Polish ancestry.
\end{tcolorbox}

\begin{tcolorbox}
\textbf{Answer:} English (9.2\%) is smaller than German (22.5\%)
\end{tcolorbox}

\caption{Case Study 1: Census-based QA under different compression schemes. Our method retains more semantic and numeric fidelity compared to other token-level approaches.}
\label{fig:casestudy1}
\end{figure*}

\begin{figure*}[htb]
\begin{tcolorbox}
\textbf{Original Prompt (NFL QA Example)}:\\
Hoping to rebound from their tough overtime road loss to the Raiders, the Jets went home for a Week 8 duel with the Kansas City Chiefs.  In the first quarter, New York took flight as QB Brett Favre completed an 18-yard TD pass to RB Leon Washington.  In the second quarter, the Chiefs tied the game as QB Tyler Thigpen completed a 19-yard TD pass to TE Tony Gonzalez.  The Jets would answer with Washington getting a 60-yard TD run.  Kansas City closed out the half as Thigpen completed an 11-yard TD pass to WR Mark Bradley. In the third quarter, the Chiefs took the lead as kicker Connor Barth nailed a 30-yard field goal, yet New York replied with RB Thomas Jones getting a 1-yard TD run.  In the fourth quarter, Kansas City got the lead again as CB Brandon Flowers returned an interception 91 yards for a touchdown.  Fortunately, the Jets pulled out the win with Favre completing the game-winning 15-yard TD pass to WR Laveranues Coles. During halftime, the Jets celebrated the 40th anniversary of their Super Bowl III championship team. \\
\textbf{Question:} How many yards was the longest TD of the game?
\end{tcolorbox}

\begin{tcolorbox}
\textbf{Compressed Prompt (LLMLingua2)}:\\
Raiders Jets Week 8 Kansas City Chiefs York Favre 18-yard Washington Chiefs Thigpen 19-yard Gonzalez 60-yard TD Kansas Thigpen 11-yard Bradley third Chiefs Barth 30-yard Jones 1-yard TD fourth Kansas Flowers touchdown Jets Favre 15-yard Coles Jets 40th Super Bowl
\end{tcolorbox}

\begin{tcolorbox}
\textbf{Compressed Prompt (QUITO)}:\\
Jets the Kansas City Chiefs. as QB Brett Favre completed to RB Leon Washington. In QB Tyler Thigpen TE Tony Gonzalez. run. Kansas WR Mark Bradley. kicker Connor Barth with RB Thomas Jones win with Favre completing to WR Laveranues Coles. During halftime, the
\end{tcolorbox}

\begin{tcolorbox}
\textbf{Compressed Prompt (Ours)}:\\
completed an 18-yard TD pass RB Tyler completed a 19-yard TD pass getting a 60-yard TD run. completed an 11-yard TD pass nailed a a 1-yard TD the 91 completing the game-winning 15-yard TD pass WR
\end{tcolorbox}

\begin{tcolorbox}
\textbf{Answer:} 91 yards (Brandon Flowers interception return)
\end{tcolorbox}
\caption{Case Study 2: Sports-related QA. Our method captures the most relevant yardage details, supporting accurate numerical reasoning.}
\label{fig:casestudy2}
\end{figure*}

These examples support our claim that while token-level compression tends to reduce syntactic completeness, our method produces more coherent and interpretable outputs than other token-level baselines, making it more suitable for applications where transparency matters.

\end{document}

%% file: definitions.tex
\usepackage{acronym}
\AtBeginDocument{%
  \providecommand\BibTeX{{%
    \normalfont B\kern-0.5em{\scshape i\kern-0.25em b}\kern-0.8em\TeX}}}

\acrodef{LLM}{large language model}
\acrodef{RAG}{Retrieval-Augmented Generation}
\acrodef{PPL}{perplexity}
\acrodef{NLP}{natural language processing}
\acrodef{ICL}{In-Context Learning}
\acrodef{SFT}{Supervised Fine-Tuning}
\acrodef{EM}{Exact Match}
\acrodef{IB}{Information Bottleneck}

%% file: figures/compare.tex
\begin{table*}[tb]
    \centering
    \setlength{\tabcolsep}{1mm} 
    \begin{tabular}{cccc}
    \toprule
        \textbf{Algorithm} & \textbf{Architecture} & \textbf{Model} & \textbf{Parameters} \\
    \midrule
         Selective Context & Transformer Decoder-Only & GPT-2 & 124M \\
         LLMLingua & Transformer Decoder-Only & Llama-2-7b & 7B \\
         LongLLMLingua & Transformer Decoder-Only & Llama-2-7b & 7B \\
         LLMLingua2 & Transformer Encoder-Only & XLM-RoBERTa-large & 355M \\
         QUITO & Transformer Decoder-Only & Qwen2-0.5b-Instruct & 500M \\
         \textbf{QUITO-X} & \textbf{Transformer Encoder-Decoder} & \textbf{FLAN-T5-small} & \textbf{80M} \\
    \bottomrule
    \end{tabular}
    \caption{Comparison of different compression algorithms in terms of architecture, model, and parameter size. Our method, based on the FLAN-T5-small model, demonstrates the effectiveness of a compact Transformer Encoder-Decoder architecture with only 80M parameters, significantly reducing computational cost while maintaining or exceeding performance compared to larger models like LLMLingua (7B) and QUITO (500M).}
    \label{tab:compare}
\end{table*}

%% file: figures/main1.tex

\begin{table*}[!ht]
    \centering
    \setlength{\tabcolsep}{1mm}
    \begin{tabular}{ccc|cccccc}
    \toprule
        dataset & model & ratio & Sel-Context & LLMLingua & LongLLMLingua & LLMLingua2 & QUITO & \textbf{QUITO-X}\\
    \midrule
    \multirow{10}{*}{\rotatebox{90}{Quoref}} & \multirow{5}{*}{\rotatebox{90}{LongChat}} 
    & 1.00 & 70.6 & 70.6 & 70.6 & 70.6 & 70.6 & 70.6 \\
     & & 0.75 & 65.3 & 46.4 & 46.5 & 65.7 & 65.6 & \textbf{68.1} \\
     & & 0.50 & 55.8 & 34.5 & 34.6 & 55.0 & 59.4 & \textbf{65.1} \\
     & & 0.25 & 40.9 & 28.2 & 28.7 & 41.5 & 52.3 & \textbf{60.8} \\
     & & 0.00 & 2.9 & 2.9 & 2.9 & 2.9 & 2.9 & 2.9  \\
      & \multirow{5}{*}{\rotatebox{90}{Llama-3}} 
      & 1.00 & 93.1 & 93.1 & 93.1 & 93.1 & 93.1 & 93.1 \\
     & & 0.75 & 90.3 & 64.9 & 65.3 & 90.7 & 89.8 & \textbf{92.6} \\
     & & 0.50 & 81.3 & 51.1 & 51.4 & 82.6 & 84.4 & \textbf{90.2} \\
     & & 0.25 & 59.3 & 43.2 & 43.3 & 65.5 & 75.8 & \textbf{86.8} \\
     & & 0.00 & 6.8 & 6.8 & 6.8 & 6.8 & 6.8 & 6.8 \\
    \midrule
    \multirow{10}{*}{\rotatebox{90}{CoQA}} & \multirow{5}{*}{\rotatebox{90}{LongChat}} 
    & 1.00 & 59.1 & 59.1 & 59.1 & 59.1 & 59.1 & 59.1 \\
     & & 0.75 & 56.6 & 44.9 & 45.4 & 57.5 & 54.6 & \underline{\textbf{59.6}} \\
     & & 0.50 & 47.0 & 36.3 & 36.4 & 50.3 & 50.4 & \underline{\textbf{59.5}} \\
     & & 0.25 & 32.1 & 30.4 & 25.9 & 41.0 & 41.4 & \textbf{55.5} \\
     & & 0.00 & 13.8 & 13.8 & 13.8 & 13.8 & 13.8 & 13.8 \\
      & \multirow{5}{*}{\rotatebox{90}{Llama-3}} 
      & 1.00 & 79.3 & 79.3 & 79.3 & 79.3 & 79.3 & 79.3 \\
     & & 0.75 & 76.5 & 62.3 & 61.8 & 74.8 & 73.1 & \underline{\textbf{79.5}} \\
     & & 0.50 & 64.1 & 50.9 & 50.4 & 69.4 & 64.6 & \textbf{78.1} \\
     & & 0.25 & 45.3 & 43.0 & 37.3 & 57.7 & 53.5 & \textbf{75.5} \\
     & & 0.00 & 18.1 & 18.1 & 18.1 & 18.1 & 18.1 & 18.1 \\
    \bottomrule
    \end{tabular}
    \caption{Experimental results of various compression methods applied at different compression ratios on the Quoref and CoQA datasets. The table shows the effectiveness of different methods, including Selective-Context, LLMLingua, LongLLMLingua, LLMLingua2, QUITO, and QUITO-X, across different compression ratios (1.00, 0.75, 0.50, 0.25, and 0.00). Our method consistently achieves the best performance at all ratios.
    }
    \label{tab:main-result}
\end{table*}

%% file: figures/main3.tex
\begin{table*}[!ht]
    \centering
    \setlength{\tabcolsep}{1mm}
    \begin{tabular}{ccc|ccccc}
    \toprule
        dataset & task & ratio & Selective-Context & QUITO & LLMLingua2 & strategy 1 & strategy 2\\
    \midrule
    \multirow{4}{*}{\rotatebox{90}{2wikimqa}} 
    & \multirow{4}{*}{Multi-Doc QA}
    & 1.00 & 55.0 & 55.0 & 55.0 & 55.0 & 55.0 \\
    & & 0.75 & 59.0 & 56.0 & 64.0 & \textbf{64.0} & 60.5 \\
    & & 0.50 & 54.5 & 58.5 & 68.0 & 67.5 & \textbf{69.0} \\
    & & 0.25 & 49.0 & 51.0 & 53.5 & \textbf{61.5} & 60.0 \\
    \midrule
    \multirow{4}{*}{\rotatebox{90}{hotpotqa}} 
    & \multirow{4}{*}{Multi-Doc QA}
    & 1.00 & 15.5 & 15.5 & 15.5 & 15.5 & 15.5 \\
    & & 0.75 & 19.0 & 21.5 & 25.5 & \textbf{31.0} & 30.0 \\
    & & 0.50 & 38.5 & 57.0 & 57.5 & \textbf{65.5} & 63.0 \\
    & & 0.25 & 46.5 & 55.0 & 52.5 & 63.0 & \textbf{69.5} \\
    \midrule
    \multirow{4}{*}{\rotatebox{90}{musique}} 
    & \multirow{4}{*}{Multi-Doc QA}
    & 1.00 & 2.5 & 2.5 & 2.5 & 2.5 & 2.5 \\
    & & 0.75 & 2.5 & 2.5 & 2.5 & \textbf{4.0} & 3.5 \\
    & & 0.50 & 10.0 & 37.0 & 40.5 & 41.5 & \textbf{43.5} \\
    & & 0.25 & 35.0 & 36.0 & 40.0 & 43.0 & \textbf{49.0} \\
    \midrule
    \multirow{4}{*}{\rotatebox{90}{Gov\_Report}} 
    & \multirow{4}{*}{Summ.}
    & 1.00 & 16.50 & 16.50 & 16.50 & 16.50 & 16.50 \\
    & & 0.75 & 16.30 & 17.44 & 17.39 & 17.72 & \textbf{17.95} \\
    & & 0.50 & 18.21 & \textbf{19.12} & 18.46 & 19.12 & 19.02 \\
    & & 0.25 & 17.96 & 19.12 & 18.04 & \textbf{19.12} & 18.90 \\
    \midrule
    \multirow{4}{*}{\rotatebox{90}{TriviaQA}} 
    & \multirow{4}{*}{Few-shot QA}
    & 1.00 & 15.0 & 15.0 & 15.0 & 15.0 & 15.0 \\
    & & 0.75 & 19.0 & 20.0 & 22.0 & \textbf{28.5} & 25.0 \\
    & & 0.50 & 27.5 & 32.5 & 22.0 & \textbf{42.0} & 38.5 \\
    & & 0.25 & 36.5 & \textbf{62.5} & 37.5 & 59.0 & 60.0 \\
    \bottomrule
    \end{tabular}
    \caption{Performance comparison across datasets under different compression ratios. We evaluate multi-doc QA, summarization, and few-shot QA tasks with Exact Match or ROUGE-L. Bold numbers indicate the best performance for each dataset and ratio combination.}
    \label{long}
\end{table*}